\documentclass[sigconf]{acmart}
\newcommand{\ind}{\perp\!\!\!\!\perp} 
\AtBeginDocument{%
  }

\setcopyright{acmlicensed}
\copyrightyear{2025}
\acmYear{2025}
\acmDOI{XXXXXXX.XXXXXXX}
\acmConference[Conference acronym 'XX]{Make sure to enter the correct
  conference title from your rights confirmation email}{June 03--05,
  2018}{Woodstock, NY}
\acmISBN{978-1-4503-XXXX-X/2018/06}

\usepackage{amsmath}
\usepackage{cancel}
\usepackage{algorithm,algpseudocode}
\usepackage{refcount}

\begin{document} 

\title{Deep Learning of Structured and Continuous Policies for Aggregated Heterogeneous Treatment Effects} 



\author{Jennifer Y. Zhang} 
\email{jenniferyt.zhang@mail.utoronto.ca} 
\affiliation{%
 \institution{University of Toronto}
 \streetaddress{27 King\'s College Cir, Toronto}
 \city{Toronto}
 \state{ON}
 \country{CA}
}

\author{Shuyang Du}
\email{shuyangdu139@gmail.com} 
\affiliation{%
 \institution{Lightspeed}
 \streetaddress{2200 Sand Hill Road}
 \city{Menlo Park}
 \state{CA}
 \country{USA}
}

\author{Will Y. Zou}
\email{will@angle.ac}
\affiliation{%
 \institution{Angle.ac}
 \city{San Francisco}
 \state{CA}
 \country{USA}
}


\begin{abstract} 
As estimation of Heterogeneous Treatment Effect (HTE) is increasingly adopted across a wide range of scientific and industrial applications, the treatment action space can naturally expand, from a binary treatment variable to a structured treatment policy. This policy may include several policy factors such as a continuous treatment intensity variable, or discrete treatment assignments. 
From first principles, we derive the formulation for incorporating multiple treatment policy variables into the functional forms of individual and average treatment effects. Building on this, we develop a methodology to directly rank subjects using aggregated HTE functions. In particular, we construct a Neural-Augmented Naive Bayes layer within a deep learning framework to incorporate an arbitrary number of factors that satisfies the Naive Bayes assumption. The factored layer is then applied with continuous treatment variables, treatment assignment, and direct ranking of aggregated treatment effect functions. Together, these algorithms build towards a generic framework for deep learning of heterogeneous treatment policies, and we show their power to improve performance with public datasets.

\end{abstract} 






\maketitle 

\section{Introduction}
\label{sec:intro} 

The direct ranking of Heterogeneous Treatment Effect~\cite{du2019improve} with many recent developments~\cite{he2024rankability, sun2024endtoendcosteffectiveincentiverecommendation, ai2024improve,kamran2024learning,zhaoaaai} offers an effective and practical framework to make machine-learning-driven business decisions for uplift marketing and cost-aware optimization. The scalable method selects subjects effectively by modeling the portfolio-wide treatment effects. In modern industrial and scientific applications, the treatment policies for subjects can be  sophisticated. For example, a subject could be treated with a discounted price which is a continuous variable, and matched with a particular product through treatment assignment. This complexity of treatments calls for algorithms that are flexible enough to handle an arbitrary number of \emph{treatment policy} variables, and one that can scale to a higher dimensionality for the action space. 

We directly address this challenge with an augmented Naive Bayes framework that can naturally incorporate many factors in a treatment policy through \emph{Bayesian decomposition}. When combined with deep learning models, this method becomes scalable and efficient without concerns for intractable or expensive partition functions in Bayesian decomposition. Not only does this method model a complex space of treatment policies, it naturally works well with the Direct Ranking approach that aggregates multiple average treatment effect functions~\cite{du2019improve}. The framework makes it possible for joint optimization of all parameters in one deep learning model. Key contributions of this paper:

\begin{itemize}
\item \textbf{Neural Augmented Bayesian Decomposition.}  We decompose the complex treatment policy factors, with each factor being ranked by a neural network model. It becomes a Neural Augmented Naive Bayes layer (NANBL) and able to handle an arbitrary number and various types of treatment variables. 
\item \textbf{Continuous Treatment Variables and Treatment Assignment.} We formulate the functional forms of individual treatment effect (ITE) and average treatment effect (ATE) with continuous treatment variables. Leveraging the NANBL framework, we formulate the model architectures and design the  non-linear forms to algorithmically enable continuous treatment variables and treatment assignment in one setting.

\item \textbf{Deep Learning of Cost-Aware Aggregated Efficiency.} Our approach aggregates multiple outcomes into a unified learning objective, allowing optimization across all parameters. The approach unifies continuous treatment variables, treatment assignments, and cost-aware aggregated treatment effects within a single deep learning framework. 

\item \textbf{Complete Causal Learning Framework with Propensity Function and Barrier Methods.} To complete the investigation, we provide the algorithms for using propensity functions with the direct ranking approach. We also propose an optimization algorithm that considers the impact of constraints. This versatile model addresses real-world needs and limitations while optimizing for market-wide efficiency.
\end{itemize} 
\section{Background} 
\label{sec:related_work} 
Previous work in user marketing and retention has explored coupon redemption~\cite{andrews2016mobile,hanna2016optimizing,Manzoor2017RUSHTT}, apology-based trust recovery~\cite{halperin2018toward,cohen2018frustration}. Churn prediction methods~\cite{vafeiadis20151,lalwani2022customer,ullah2019churn} use boosting, SVM, and classification techniques to directly predict user behavior. However, these methods primarily focus on outcome prediction or estimating average treatment effects, with limited attention to user-level selection or optimization under constraints.

Rubin's causal inference framework~\cite{rubin1974estimating} provides a foundation for studying treatment effects. User instances are treated with specific actions, and the observed outcomes are used to fit models. A significant approach involves statistical methods such as meta-learners~\cite{kunzel2017meta},  decomposing the learning algorithms into composite models. Another approach uses decision trees and random forests~\cite{chen2016xgboost}, such as uplift trees\cite{rzepakowski2012decision}, causal  forests~\cite{wager2017estimation, athey2016recursive}, boosting~\cite{powers2017some}, and latent variable models~\cite{louizos2017causal}, to build powerful causal inference models. Quasi-oracle estimation~\cite{nie2017quasi} has emerged as a framework for learning heterogeneous treatment effects. It effectively estimates treatment effects for single outcomes by considering both \emph{Conditional Treatment Effect (CTE)} and \emph{Average Treatment Effect (ATE)}. These algorithms are limited to single-outcome scenarios and cannot handle multiple outcomes or benefit-cost trade-offs. 

While the above models have shown strong performance in estimating conditional treatment effects, their modular structures, consisting of independently trained components, limit end-to-end optimization. These modular components are learned separately and greedily on the training data. Moreover, R-learner and Meta-learner models focus on the treatment effect of a single outcome or a single metric uplift, instead of an objected aggregated with multiple uplift functions. 

Recent work explores more sophisticated causal learning and uplift modeling techniques tailored to real-world problems. \cite{du2019improve} presents the first method to aggregate multiple treatment effect outcomes into a comprehensive treatment effect measure that can be jointly optimized. Recently, \cite{he2024rankability} introduces a rankability-enhanced framework that produced recent performance benchmarks. This work optimizes revenue uplift with multiple additional objectives and offers improvements in handling heavy tail distributions. \cite{kamran2024learning} discusses treatment allocation under resource constraints from a ranking perspective. \cite{sun2024endtoendcosteffectiveincentiverecommendation, zhaoaaai} offer ways to inject domain knowledge and budget allocation improvements.  Till now, most approaches in scientific or industrial uses of heterogeneous treatment effects focus only on ranking with one treatment variable, often with a binary treatment variable. 

For the use of continuous variables as treatment, \cite{imai2004causal} offers the first introduction of the topic with discussions on generalizing the propensity score. \cite{chernozhukov2019semi} is a recent work for continuous off-policy evaluation with a theoretical focus not yet applied to HTE domains. \cite{kallus2018policy} discusses continuous treatments for off-policy evaluation on singular rewards with experiments on simulated data. \cite{kreif2015evaluation} presents a scientific application of the generalized propensity score in traumatic brain injury. Although continuous variable treatment has been explored, there lacks of a framework that can handle complex treatment policies with several continuous, binary, discrete treatment variables. Further, a scalable approach hasn't been proposed, one that can be applied to real-world datasets and scalable in cost-aware applications with aggregated treatment effect functions. 

To address the lack of studies for complex treatment policies, we propose a Neural Augmented Naive Bayes framework that supports arbitrary decomposition of treatment variables beyond the one treatment variable or binary treatment settings. Moreover, our algorithm not only predicts treatment effects but also combines multiple outcomes into comprehensive effectiveness measures that can be jointly optimized under constraints. Our end-to-end deep learning optimization framework unifies continuous treatment variables, treatment assignments, and cost-aware aggregated treatment effects, expanding the scope of uplift modeling. 
\section{Algorithms} 
\label{sec:algorithms} 
We develop the deep learning approach in this section. First, we derive the changes in the treatment effect functional forms ($\tau$) for continuous treatments; then, we introduce the Neural Augmented Naive Bayes layer (NANBL); next, we integrate it with deep learning on aggregation of multiple treatment effect functions ($\tau$); we then explain how to work with the treatment assignment task by stacking the NANBL and how to incorporate treatment factors; finally, we provide insights about incorporating propensity and constraints.

\subsection{Functional Forms for Treatment Effects with Continuous Treatment}
\textbf{Individual Treatment Effect (ITE).} ITE is the difference between the potential outcomes under treatment ($Y_1$) and control ($Y_0$) for a given individual. With binary treatment, we evaluate cases where outcome values are available. For example, in the training set where the data is labeled, we can directly compute the treatment effect: $\tau = Y_1 - Y_0$. When the outcome needs to be predicted, i.e. we only have covariates ($\mathbf{x}$), we use a function for prediction: $\tau(\mathbf{x}): \mathbb{R}^n \to \mathbb{R}$, where the output is a real scalar. Given $\mathbf{x} = \overline{\mathbf{x}}$ is observed, the treatment function maps from $\overline{\mathbf{x}} \in \mathbb{R}^n$ to $\mathbb{R}$.

In the case of continuous treatment variables, we argue that the functional forms will be generalized to accommodate a broader range of input values. The estimates are the same for the cases where the outcome values and continuous treatment values $\rho_c$ are available. Specifically, the treatment effect conditioned on $\rho_c$ is given by: $\tau(\mathbf{x} \mid \rho_c) = Y_{\rho_c} - Y_0 \mid \rho_c$,  where $Y_{\rho_c}$ is the output observed when the continuous treatment $\rho_c$ is applied and $Y_0$ is the baseline outcome. When the outcome is to be predicted, the model is derived from: $\tau(\mathbf{x} \mid \rho_c): \mathbb{R}^n \times \mathbb{R} \to \mathbb{R}$. Given $\mathbf{x}=\overline{\mathbf{x}}$ is observed, the output is a function that takes in the continuous treatment variable $\rho_c$ as an argument:  
\begin{align*}
    f(\rho_c)=\tau(\overline{\mathbf{x}}\mid\rho_c):\mathbb{R} \to\mathbb{R}.
\end{align*}
This expression, given $\mathbf{x}$ is observed, produces an output given any value of the continuous treatment variable $\rho_c$ in its scope.

\textbf{Average Treatment Effect (ATE)}. In the binary treatment case, the ATE can be written as an expected value: 
\begin{align*}
\tau^*(\mathbf{x}) = \mathbb{E}[Y_1 - Y_0 \mid \mathbf{x}] = \mathbb{E}[Y_1 \mid \mathbf{x}] -  \mathbb{E}[Y_0 \mid\mathbf{x}].
\end{align*}
In the case of continuous treatment, assuming there is \textit{a controlled cohort of subjects where the treatment was not given ($\rho_c=0$)}, the ATE can be written as: 
\begin{align}
\label{eq:functional_form_ate_cont_base}
\tau^*(\mathbf{x} \mid \rho_c) = \mathbb{E}[Y_{\rho_c} - Y_0 \mid \mathbf{x} ] = \mathbb{E}[Y_{\rho_c} \mid \mathbf{x} ] - \mathbb{E}[Y_0 \mid \mathbf{x}].
\end{align}


\subsection{Neural Augmented Naive Bayes Layer} 
In order to estimate the functional form of $ \mathbb{E}[Y_{\rho_c} \mid\mathbf{x}]$ in Eq. ~\ref{eq:functional_form_ate_cont_base}, we introduce the Neural Augmented Naive Bayes layer. 

The method of using neural networks to augment Naive Bayes and Markov Chain algorithms ~\cite{azeraf2021improving} has been studied. However,
they have not been applied to heterogeneous treatment effects or causal learning. 
We expand $ \mathbb{E}[Y_{\rho_c} \mid \mathbf{x}]$ from Eq. ~\ref{eq:functional_form_ate_cont_base} in the case where $\rho_c$ is observed, such as in a case where the training set has treatment and controlled cohorts: 
\begin{align}
\label{eq:poterior_expansion_of_ate_first_term}
    \mathbb{E}[Y_{\rho_c} \mid \mathbf{x}] = \sum_{\mathbf{x} \in \text{cohort}, \rho_c \neq 0} p(\mathbf{I}_\mathbf{x} \mid \rho_c, \mathbf{x}) Y_{\rho_c}.
\end{align}
\begin{figure}
  \centering \includegraphics[width=0.175\textwidth]{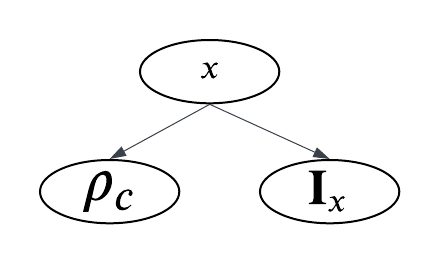} 
  \caption{Bayesian Network across covariates, the continuous treatment variable, and the treatment selection indicator.} 
  \label{fig:hcl_cont_naive_bayes} 
\end{figure} 

Here, we consider the probability $p(\mathbf{I}_\mathbf{x} \mid \rho_c, \mathbf{x})$ as the posterior probability of the subject being chosen for treatment with observed $\rho_c$ and covariates $\mathbf{x}$ in the treatment cohort. The variable $\mathbf{I}_\mathbf{x}$ is used as an effectiveness measure for treating subjects, e.g. for maximum uplift or maximizing an aggregated uplift objective. Eq.~\ref{eq:poterior_expansion_of_ate_first_term} approximates the conditional expected outcome under a continuous treatment by aggregating observed outcomes in the training data, weighted by the posterior probability. 

To construct a model for direct ranking of heterogeneous treatment effects under continuous treatments, we integrate multiple sub-models in a unified framework that supports joint optimization of model parameters. We observe the posterior probability in  Eq.~\ref{eq:poterior_expansion_of_ate_first_term} can be transformed using the Bayes Rule:
\begin{align}
\label{eq:bayesian_decomposition_cont}
p(\mathbf{I}_\mathbf{x} \mid\rho_c, \mathbf{x}) = \frac{p(\rho_c \mid \mathbf{I}_\mathbf{x}, \mathbf{x})p(\mathbf{I}_\mathbf{x} \mid \mathbf{x})}{\sum_{\mathbf{x}\in \text{cohort}, \rho_c \neq 0}p(\rho_c \mid \mathbf{I}_\mathbf{x}, \mathbf{x})p(\mathbf{I}_\mathbf{x} \mid \mathbf{x})}.
\end{align}

We make the Naive Bayes assumption across the variable $\mathbf{I}_\mathbf{x}$ and the variable $\rho_c$, so that $ \mathbf{I}_\mathbf{x} \ind \rho_c \mid \mathbf{x}$, and it follows that $p(\rho_c|\mathbf{I}_\mathbf{x}, \mathbf{x}) = p(\rho_c\mid \mathbf{x})$. For Neural Augmented Naive Bayes, we represent the likelihood $p(\rho_c \mid \mathbf{x})$ and prior $p(\mathbf{I}_\mathbf{x}\mid\mathbf{x})$ with functional forms $f$ and $g$:
\begin{align*}
    p(\mathbf{I}_\mathbf{x} \mid \mathbf{x}) &= f(\mathbf{x}), \\
    p(\rho_c \mid \mathbf{x}) &= g(\mathbf{x}, \rho_c) \\
    &= \sigma(\rho_c - \hat{g}(\mathbf{x}))(1 - \sigma(\rho_c - \hat{g}(\mathbf{x}))) \\
    &= \text{bell}(\rho_c - \hat{g}(\mathbf{x})),
\end{align*}
where $f(\mathbf{x})$ and $\hat{g}(\mathbf{x})$ denote forward functions of neural networks or multi-layer perceptron models, $\sigma$ denotes the sigmoid function with output in $(0, 1)$. The network $f$ ends with a sigmoid non-linearity. $g$ takes the form of the derivative of the sigmoid function, forming a bell curve centered around  $\hat{g}(\mathbf{x})$ that predicts the center-value of the distribution. The resulting $f(\mathbf{x})$ and $g(\mathbf{x},\rho_c)$ both give strictly positive outputs. Eq.~\ref{eq:poterior_expansion_of_ate_first_term} becomes: 
\begin{align}
\label{eq:nanbl_layer}
p(\mathbf{I}_\mathbf{x} \mid \rho_c, \mathbf{x}) = \frac{g(\mathbf{x}, \rho_c)f(\mathbf{x})}{\sum_{\mathbf{x}\in \text{cohort}, \rho_c \neq 0}g(\mathbf{x}, \rho_c)f(\mathbf{x})}.
\end{align}

Because the denominator is consistently larger than 0, this expression is differentiable with respect to $f$, $g$, $\hat{g}$ and $\mathbf{x}$. We call this the \textbf{Neural Augmented Naive Bayes Layer} \emph{(NANBL)}. 

The advantage of a NANBL layer is computational efficiency. Instead of computing partition functions which is challenging for probability distributions, we apply computationally efficient deep learning models to express prior and likelihood functions so the partition function can be easily computed. Furthermore, deep learning architectures such as softmax and normalization make it efficient to express the NANBL as a component layer, making it scalable and easily integrated into deeper architectures. 

\subsection{Treatment Assignment and Recursive Stacking of NANBL for Treatment Policies} 
We extend the model to deal with treatment assignment. It is the discrete choice variable denoted as $t_a$, or treatment matching. This function enables the model to handle the assignment of discrete treatments, e.g. which medicine to offer to the patient, or which product to market to a user. We decompose the posterior as follows: 
\begin{align}
\label{eq:treatment_assignment_decomposition}
p(\mathbf{I}_\mathbf{x} \mid t_a, \rho_c, \mathbf{x}) &= \frac{p(t_a \mid \mathbf{I}_\mathbf{x}, \rho_c, \mathbf{x})p(\mathbf{I}_\mathbf{x} \mid \rho_c, \mathbf{x})}{\sum_{\mathbf{x}\in D, \rho_c \neq 0}p(t_a \mid \mathbf{I}_\mathbf{x}, \rho_c, \mathbf{x})p(\mathbf{I}_\mathbf{x} | \rho_c, \mathbf{x})}.
\end{align}

\begin{figure}[ht]
  \centering \includegraphics[width=0.40\linewidth]{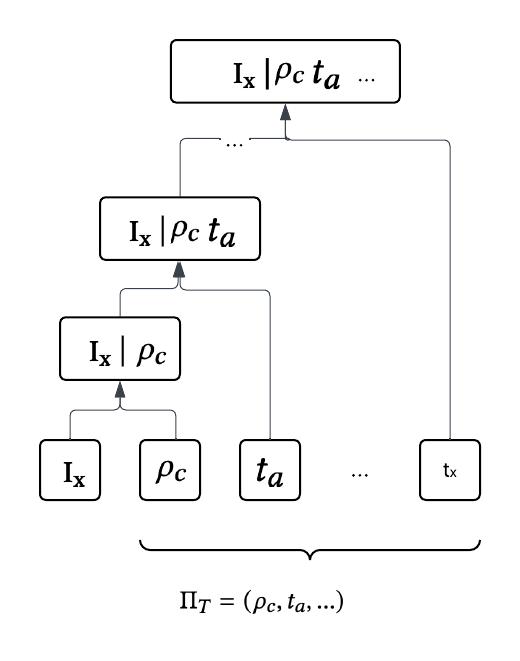} 
  \caption{Recursively applying the NANBLs for an arbitrary number of factors.} \label{fig:recursion_nanbl}
\end{figure}

As shown above, $p(\mathbf{I}_\mathbf{x} \mid \rho_c, \mathbf{x})$ is expressed as the output of a neural network, and we can use a new network to represent the new likelihood $p(t_a \mid \mathbf{I}_\mathbf{x}, \rho_c, \mathbf{x})=f_{t_a}(\mathbf{x})$ under the Naive Bayes assumption. We can thus \emph{recursively apply Bayesian decomposition} to construct another NANBL layer to arrive at an expression analogous to Eq.~\ref{eq:nanbl_layer} to include an additional treatment factor. This process can be repeated as needed. As shown in Fig.~\ref{fig:recursion_nanbl}, we \emph{recursively} construct the network as additional factors are introduced into the model. At each recursive step, the output of the neural network from the previous step is used for computing the prior in the next decomposition step. This structure enables the model to flexibly integrate multiple variables into a coherent probabilistic inference framework. We define this as the methodology for ~\emph{stacking} the NANBLs. 

In fact, our model can handle an arbitrary number of treatment factors as long as the Naive Bayes assumption is applied. The set of treatment factors, such as continuous intensity and treatment assignment, composes the \emph{treatment policy} $\Pi$: 
\begin{align*}
    \Pi_\mathbf{x} = (\rho_c, t_a, ...).
\end{align*}
\begin{algorithm}
    \caption{Recursive Forward Function for Stacked NANBL Layers}
    \label{alg:nanbl_recursion}
    \begin{algorithmic}[1] 
        \Procedure{RecursiveFwd}{$Set_{\text{var}}$, $cohort$} 
        \If{$|Set_{\text{var}}|$ = 1} 
        \State $v \gets Set_{\text{var}}.last$
        \State $f \gets ForwardFunction(v)$ \Comment{Get fwd function}
        \State $\mathbf{x} \gets Covariates(v)$ \Comment{Get features}
        \State \textbf{return} $f(x)$
        \Else \Comment{If Set has more than one element}
        \State $v \gets Set_{\text{var}}.last$
        \State $S \gets Set_{\text{var}}\setminus v$
        \State $f \gets ForwardFunction(v)$ 
        \State $\mathbf{x} \gets Covariates(v)$
        \State $\mathbf{p} = f(x)$ \Comment{Return tensor of all users}
        \State $\mathbf{l} = \text{RecursiveFwd}(S, cohort)$ \Comment{Recursive call}
        \State $\mathbf{d} = \mathbf{l}\cdot\mathbf{p}$   \Comment{Dot product}     
        \State \textbf{return} $\frac{\mathbf{d}}{\sum_{cohort} \mathbf{d}}$ \Comment{Normalize w. partition func.}
        \EndIf 
        \EndProcedure
    \end{algorithmic}
\end{algorithm}

Alg.~\ref{alg:nanbl_recursion} explains how to construct a stacked NANBL model given a set of treatment variables $Set_{\text{var}}$,  their corresponding covariates in \textit{cohort}, and their corresponding neural network forward functions. The forward function \textit{ForwardFunction} defined in Alg.~\ref{alg:nanbl_recursion} is sufficient~\footnote{We can assume the process to obtain gradients through back-propagation is automatically derived and implemented.}.

The stacked NANBLs is a powerful way to handle arbitrary \emph{treatment policies} with a large number of and various types of treatment variables. 

\subsection{Structured and Continuous Policy Model (SCPM) for HTE} 
\label{sec:SCPM}



With the above NANBL formulation for posterior probabilities, we build towards training a deep learning model that is flexible with combining average treatment effect functions. Our work builds on the direct ranking work to aggregate multiple treatment effect functions~\cite{du2019improve, he2024rankability}. 

\textbf{Wholistic Optimization.} In our aggregated treatment effect experiments, we apply a cost-aware optimization objective that is suitable for user engagement or marketing:
\begin{align} 
\label{eq:pstatement_unconstrained_opt} 
\text{maximize} \quad \frac{\text{Incremental Value}}{\text{Incremental Cost}} = \frac{\tau^{*r}(\Pi_\mathbf{x}, \mathbf{\theta}, \mathbf{x})}{\tau^{*c}(\Pi_\mathbf{x}, \mathbf{\theta}, \mathbf{x})},
\end{align} 
where $\theta$ is the model parameter.

The objective function is an \emph{aggregation of treatment effect functions}. Maximizing it requires treatment effect functional forms of model parameters and the input, namely, $\tau (\Pi_\mathbf{x}, \mathbf{\theta}, \mathbf{x})$. 

\textbf{Functional form for average treatment effect.} The average treatment effect function is an expected value of the outcome $Y$. Te $\tau$ function can be expressed with effectiveness probabilities and the observed outcomes Y (users indexed with $i$ and covariates $\mathbf{x}$): 
\begin{align}
\label{eq:treatment_effect_functional_form}
&\tau(\Pi_\mathbf{x}, \mathbf{\theta}, \mathbf{x}) \notag = \mathbf{E}(Y_1-Y_0|\mathbf{x}, \mathbf{\theta}) \notag\\
&=\sum_{T_i=1} p_i(I_\mathbf{x}\mid\Pi_\mathbf{x},\mathbf{\theta}, \mathbf{x}) Y^{(i)} - \sum_{T_i=0} p_i(I_\mathbf{x}\mid\Pi_\mathbf{x},\mathbf{\theta}, \mathbf{x}) Y^{(i)}.
\end{align} 

The above form of $\tau (\Pi_\mathbf{x}, \mathbf{\theta}, \mathbf{x})$ is expressed in terms of user effectiveness probabilities $p_i(I_\mathbf{x}\mid\Pi_\mathbf{x},\mathbf{\theta}, \mathbf{x})$. 

\textbf{Normalized Effectiveness Probabilities.} We leverage the NANBL layer (Eq.~\ref{eq:bayesian_decomposition_cont} and Eq.~\ref{eq:nanbl_layer}) in the previous section, which provides probabilities normalized by the partition function. Wherever required, a softmax function can also be applied across the subject cohorts to normalize model scores ~\cite{shazeer2017outrageously, brown2020language,huang2013learning,valizadegan2009learning}: 
\begin{align}
\label{eq:softmax_effectiveness_probabilities}
p(I_\mathbf{x}\mid\Pi_\mathbf{x}, \mathbf{\theta}, \mathbf{x}) &= \text{softmax}(s)=\frac{\exp(s)}{\sum_{\text{cohort}} \exp(s)},
\end{align}
where the scores $s$ can be the output by the NANBL layer. With normalized probabilities, $\tau(\Pi_\mathbf{x}, \mathbf{\theta}, \mathbf{x})$ can be expressed and aggregated in Eq.~\ref{eq:pstatement_unconstrained_opt} to form the ROI maximization objective. 

\textbf{Optimizing with Deep Learning}. Putting together Eq.~\ref{eq:pstatement_unconstrained_opt}, Eq.~\ref{eq:treatment_effect_functional_form}, and Eq.~\ref{eq:softmax_effectiveness_probabilities}, and  the learning objective is given in Eq.~\ref{eq:drm_deep_learning_summary}, which includes a regularization term.
\begin{align}
  \label{eq:drm_deep_learning_summary}
  \hat{\theta} = \text{argmax}_{\theta}\left \{ \frac{\tau^{*r}(\Pi_\mathbf{x}, \mathbf{\theta}, \mathbf{x})}{\tau^{*c}(\Pi_\mathbf{x}, \mathbf{\theta}, \mathbf{x})}-\Lambda_n(\cdot )  \right \}.
\end{align}

The model parameters could be trained by maximizing the objective function.


\subsection{Propensity Weighting}
For experiments where treatment is not given randomly, propensity~\cite{rubin1974estimating, nie2017quasi} ensures proper weighting is given to the model.  In our algorithm, the propensity function $e(\mathbf{x})$ can be estimated using a separate regression model. We derive the full form of propensity weighting for  Direct Ranking~\cite{du2019improve} in Appendix~\ref{apB}. We offer the simplified form of the treatment effect function and its relationship with the propensity function for user $i$:
\begin{align}
\label{eq:treament_effect_function_with_propensity}
\tau^{*} = \hat{e} \sum\limits_{i=1}^{n} \frac{1}{e(\mathbf{x}^{(i)})} Y^{(i)} p_i \mathbb{I}_{T_i=1} - (1-\hat{e}) \sum\limits_{i=1}^{n} \frac{1}{1-e(\mathbf{x}^{(i)})} Y^{(i)} p_i \mathbb{I}_{T_i=0},
\end{align}
where $\hat{e}$ is the number of treated instances over the total number of instances. 

The propensity-weighted objective is built using Eq.~\ref{eq:treament_effect_function_with_propensity} combined into Eq.~\ref{eq:drm_deep_learning_summary}. In experiments, we find Direct Ranking Model with propensity weighting is useful for interpreting counterfactuals, discussed in Section~\ref{sec:empirical_results}\footnote{To apply propensity in DRM, we note that the training and prediction data assumptions as well as their measurement metrics should align.}.

\subsection{Incorporating Constraints} 
To handle constraints, we apply the barrier method~\cite{norcedal2006numerical}. This can help address cases where we act on a percentage of subjects.

\textbf{Barrier Function}. We define barrier functions to introduce penalties if the desired constraint is not met. The objective with barrier function can be computed given any batch of data during optimization. The method works with either a fixed treatment percentage $P$  (\emph{Percentage Constraint} problem) or a fixed cost budget $B$ (\emph{Cost Budget Constraint} problem).


The barrier method can be seen as a form of dynamic pooling~\cite{jarrett2009best}. The model can be trained with gradient methods. It dynamically creates connection patterns in the neural network to focus on the largest activations. See derivations and more algorithm implementation in Appendix \ref{constraint-ranking}.


\section{Experiments} 
\label{sec:empirical_results} 

\subsection{Evaluation Methodology}
For single $\tau$ objective modeling, we adopt the evaluation methodology in the most recent HTE ranking work~\cite{he2024rankability} and compare it with the state-of-the-art. For the modeling of aggregated $\tau$ functions, we adopt the method in the classic work~\cite{du2019improve}.

\textbf{Area Under Uplift Curve (AUUC)}. The uplift curve ranks individual samples descendingly according to predicted uplift $\tau^r$ (in X-axis) and the cumulative sum of observed uplift (in Y-axis) \cite{betlei2021}. We follow the implementation in \cite{he2024rankability} for calculating the area under this curve as AUUC.

\textbf{Area Under QINI Curve (AUQC)}. The QINI curve computes the difference in cumulative outcomes between treated and control groups over increasing percentiles of the population sorted by predicted uplift. It captures how effective the model is at prioritizing individuals who will benefit most from treatment. We implement AUQC as an evaluation metric following \cite{he2024rankability}.

\textbf{Kendall Rank Correlation Coefficient (KRCC)}. KRCC measures the correlation between two ranked lists—in our case, the predicted uplift rankings and an approximation of the true uplift rankings. The whole dataset is split into buckets and we use the treatment and control group data in each bucket to approximate the true uplift effect \cite{he2024rankability}.

\textbf{LIFT@h}. This metric measures the difference between the mean response of treated individuals and that of controlled individuals in the top $h=30$ percentile of all individuals ranked by the uplift model \cite{he2024rankability}.



\textbf{Area Under Cost Curve (AUCC)}. We follow the evaluation method given by~\cite{du2019improve}. To evaluate ranking of aggregated $\tau$ functions, we plot the cumulative incremental value $\tau^{r}$ against incremental cost $\tau^{c}$. This is called the \textbf{\emph{Cost Curve}}. Similar to the Area Under Curve of ROC curve, we define the normalized area under cost curve as the area under curve divided by the area of rectangle extended by maximum incremental value and cost. 

\subsection{Datasets} 
\textbf{Ponpare}~\cite{coupon-purchase-prediction}. We apply careful and extensive data pre-processing to ensure data quality, and intend to makes Ponpare dataset a valuable benchmark for models with structured and continuous treatment policies.  We share the processing code with the research community. The dataset contains user features, coupon characteristics, and user-coupon interaction. The discrete choice features, such as genre and area name are processed into sparse one-hot vectors. The dense float features are normalized \footnote{Normalized by subtracting mean and dividing by the standard deviation.}. After processing, the user and coupon feature have dimensions of $50$ and $160$, respectively. We use the intended `training' set as the full dataset, since the `test' set is not made public. The full data has $2517206$ samples. We subsample by $5$ for our experiments with $839069$ samples. We use a $3/1/1$ split for training, validation and test set. The treatment cohort is segmented by whether a user receives a coupon with a discount above the median of all coupons. The \textbf{continuous treatment intensity} captures both magnitude of discount and user engagement with that discount (through item count). The gain dimension is the number of units purchased for $\tau^r$. The cost dimension is the number of items sold multiplied by the difference between discount price and the original price for $\tau^c$. 

\textbf{US Census 1990}~\cite{uscensulink}. Each sample in this dataset contains personal features such as native language and education. We select people with one or more children (\emph{`iFertil'} $\geq1.5$), born in the U.S. (\emph{`iCitizen'} = 0), and less than $50$ years old (\emph{`dAge'} $<$ 5), resulting in a dataset with $225814$ samples and input dimension of $46$. \footnote{\label{fourth}Please see Appendix~\ref{appendix:data_sets} for more details to process the datasets.}

\textbf{Covertype}~\cite{covertype_31}. The Covertype Dataset contains the cover type of northern Colorado forest areas with tree classes, distance to hydrology, distance to wild fire ignition points, elevation, slope, aspect, and soil type. After processing, the data has $244365$ samples and $51$ features for model input. \footnotemark[\getrefnumber{fourth}]

\subsection{Experiment Results and Analysis} 
We compare our structured and continuous policy model (SCPM) algorithm with DragonNet \cite{Shi2019} and RERUM DragonNet \cite{he2024rankability}, CFR \cite{shalit2017estimatingindividualtreatmenteffect} and RERUM CFR \cite{he2024rankability} on the Ponpare dataset. For comparison with these state-of-the-art models, we adopt the same \emph{single treatment effect function $\tau$} approach as~\cite{he2024rankability}. For aggregated treatment effect functions, we show the effectiveness of aggregating multiple $\tau$ functions, by comparing with Direct Ranking Model (DRM) \cite{du2019improve}, the classic R-learner~\cite{nie2017quasi}, Duality R-learner \footnote{See Appendix \ref{dual_r_learner} for detailed description of Duality R-Learner.}, and show the performance of Constraint Ranking Model on USCensus and Covtype datasets. We also provide analysis and discussions of SCPM and DRM with propensity.~\footnote{See detailed model implementations and source code in Appendix \ref{model_config}.}

\begin{table*}[ht]
  \caption{Summary of benchmark results on Public Datasets.} 
  \label{tab:summary_result_table}
  \begin{tabular}{llllllllll}
    \toprule
    Ponpare Single $\tau$ & AUUC & AUQC & KRCC & LIFT@30 & Aggregated $\frac{\tau^r}{\tau^c}$ AUCC & USCensus & Covtype \\
    \midrule 
    DragonNet & 0.5512$\pm$ 0.0296 & 0.5540$\pm$0.0295 & 0.0560$\pm$0.0574 & 0.0116$\pm$0.0032 & Random & 0.500 & 0.500\\
    CFR (mmd) & 0.6160$\pm$0.0125 & 0.6282$\pm$0.0092 & 0.1726$\pm$0.0354 & 0.0134$\pm$0.0018 & R-learner G & 0.529 & 0.817\\
    CFR (was) & 0.6149$\pm$0.0140 & 0.6298$\pm$0.0122 & 0.1638$\pm$0.0282 & 0.0125$\pm$0.0017 & R-Learner MLP & 0.506 & 0.651\\
    RERUM DragonNet & 0.5186$\pm$0.0510 & 0.5218$\pm$0.0507 & 0.0273$\pm$0.0704 & 0.0124$\pm$0.0017  & Duality R-learner & 0.545 & 0.825\\
    RERUM CFR (mmd) & 0.6160$\pm$0.0125 & 0.6282$\pm$0.0092 & 0.1726$\pm$0.0354 & 0.0134$\pm$0.0018  & Causal Forest & 0.499 & 0.498 \\
    RERUM CFR (was) & 0.6454$\pm$0.0230 & 0.6514$\pm$0.0180 & \textbf{0.2144}$\pm$0.0165 & 0.0148$\pm$0.0022  & Direct Ranking & 0.604 &\textbf{0.907} \\
    SCPM & \textbf{0.6559}$\pm$0.0059 & \textbf{0.6522}$\pm$0.0066 & 0.2098$\pm$0.0443 & \textbf{0.0185}$\pm$0.0014 & Constrained Ranking & \textbf{0.715} & 0.770 \\
    \bottomrule
  \end{tabular}
\end{table*}
\vspace{-0.3mm}


\textbf{Overall SCPM Performance.}
Compared with state-of-the-art, SCPM as a combined deep learning framework significantly out-performs DragonNet, CRF, and RERUM~\cite{he2024rankability} improved versions of DragonNet and CRF. Note the test set ranking is only performed using the \emph{prior} neural network in the NANBL layer, $f(\mathbf{x})$, or $p(\mathbf{I_x} \mid \mathbf{x})$ in Eq.~\ref{eq:poterior_expansion_of_ate_first_term}. Even though the training is performed on Ponpare with coupon features for treatment assignment and continuous treatment variables, those cardinalities of data were \emph{not} used during test time. The model has generalized the learning from coupon features and continuous treatment variables, and is able to perform better at test time even though only the basic ranker on user features is used. The results are given in Table \ref{tab:summary_result_table}, showing SCPM consistently performs on three out of four evaluation metrics. 

\begin{figure}[htbp]
    \centering
    \includegraphics[width=0.75\linewidth]{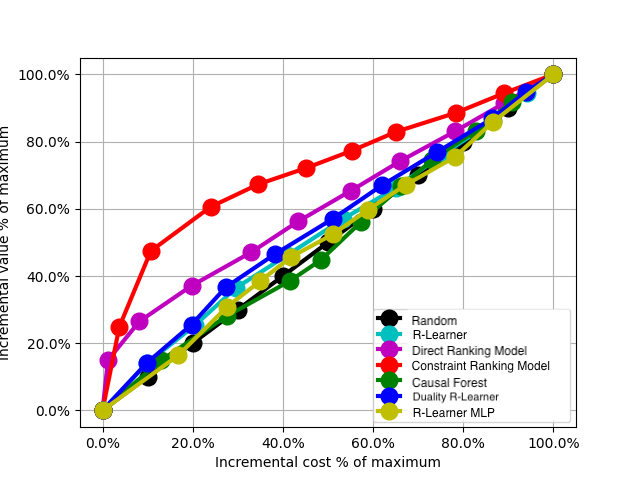} 
    \caption{Cost curve for US Census dataset.}
    \label{fig:uscensus_result}     
\end{figure}

\begin{figure}[ht]
    \centering
    \includegraphics[width=0.75\linewidth]{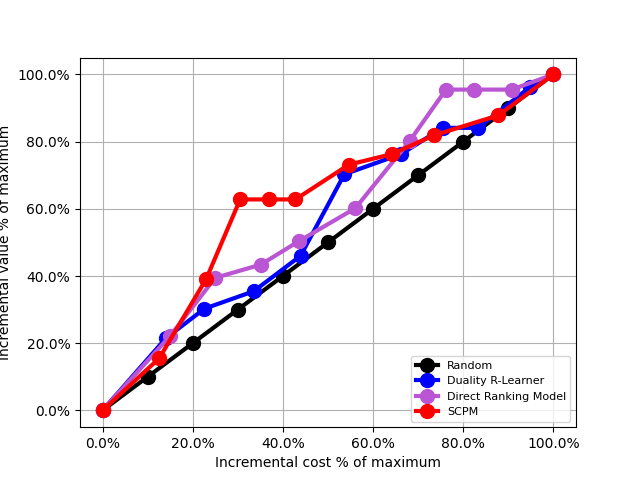}
    \caption{Cost curve for SCPM on Ponpare dataset.}
    \label{fig:scpm_vs_drm}
\end{figure}

\textbf{Significance of Aggregated Treatment Effect.} Table~\ref{tab:summary_result_table} shows Constrained Ranking and Direct Ranking Model perform consistently better with the AUCC metric than R-learner and Duality R-Learner, often by more than \emph{10\%}. Fig. \ref{fig:uscensus_result} also illustrates the cost curve across models. We attribute this performance to the models' ability to aggregate multiple treatment effect functions into a combined objective, and optimize all the parameters jointly. We show in Fig. \ref{fig:scpm_vs_drm}, that SCPM with an objective with aggregated $\tau$ functions, performs better than Direct Ranking and duality R-Learner. On Ponpare dataset, SCPM has an AUCC of $ 0.608$, compared with DRM  of $ 0.578$, and Duality R-learner of $0.558$.

\vspace{-0.3mm}
\vspace{-0.3mm}
\textbf{Model Generalization with Propensity.} We validate that model generalization on public datasets can be improved with propensity weighting. We compute the use propensity weighted objective (Eq.~\ref{eq:drm_deep_learning_summary}, Eq.~\ref{eq:treament_effect_function_with_propensity}) on the test set. Our experiments in Table~\ref{tab:generalization_propensity_table} show that incorporating propensity function significantly enhances the test metric of both Direct Ranking and R-Learner. 

\begin{table}[ht]
\caption{Test set generalization scores with propensity across different treatment percentages.}
    \centering
    \resizebox{\columnwidth}{!}{
        \begin{tabular}{|c|c|c|c|c|}
            \hline
            \multicolumn{5}{|c|}{\textbf{US Census: test set generalization scores}} \\
            \hline
            $q$ & R-Learner & R-Learner w. Propensity & DRM & DRM w. Propensity \\
            \hline
            20\% & 0.0234 & \textbf{0.0370} & 0.4426 & 0.3896 \\
            \hline
            40\% & 0.0149 & \textbf{0.0271} & 0.3148& \textbf{0.5151} \\
            \hline
            60\% & 0.0080 & \textbf{0.0129} & 0.3679 & \textbf{0.5684} \\
            \hline
            80\% & 0.0053 & \textbf{0.0099} & 0.4224 & \textbf{0.6117} \\
            \hline
            100\% & 0.0039 & \textbf{0.0072} & 0.5144 & \textbf{0.8085} \\
            \hline
            \multicolumn{5}{|c|}{\textbf{Covtype: test set generalization scores}} \\
            \hline
            q & R-Learner & R-Learner w. Propensity & DRM & DRM w. Propensity \\
            \hline
            20\% & 0.0180 & \textbf{0.0203} & 0.1447 & \textbf{0.6034} \\
            \hline
            40\% & 0.0086 & \textbf{0.0105} & 0.1543 & \textbf{0.6747} \\
            \hline
            60\% & 0.0056 & \textbf{0.0076} & 0.1610 & \textbf{0.7442} \\
            \hline
            80\% & 0.0043 & \textbf{0.0062} & 0.1148 & \textbf{0.8749} \\
            \hline
            100\% & 0.0035 & \textbf{0.0049} & 0.0810 & \textbf{1.1420} \\
            \hline            
        \end{tabular}
    }
    \label{tab:generalization_propensity_table}
\end{table}

\begin{figure}[ht] 
  \centering 
  \includegraphics[width=0.70\linewidth]{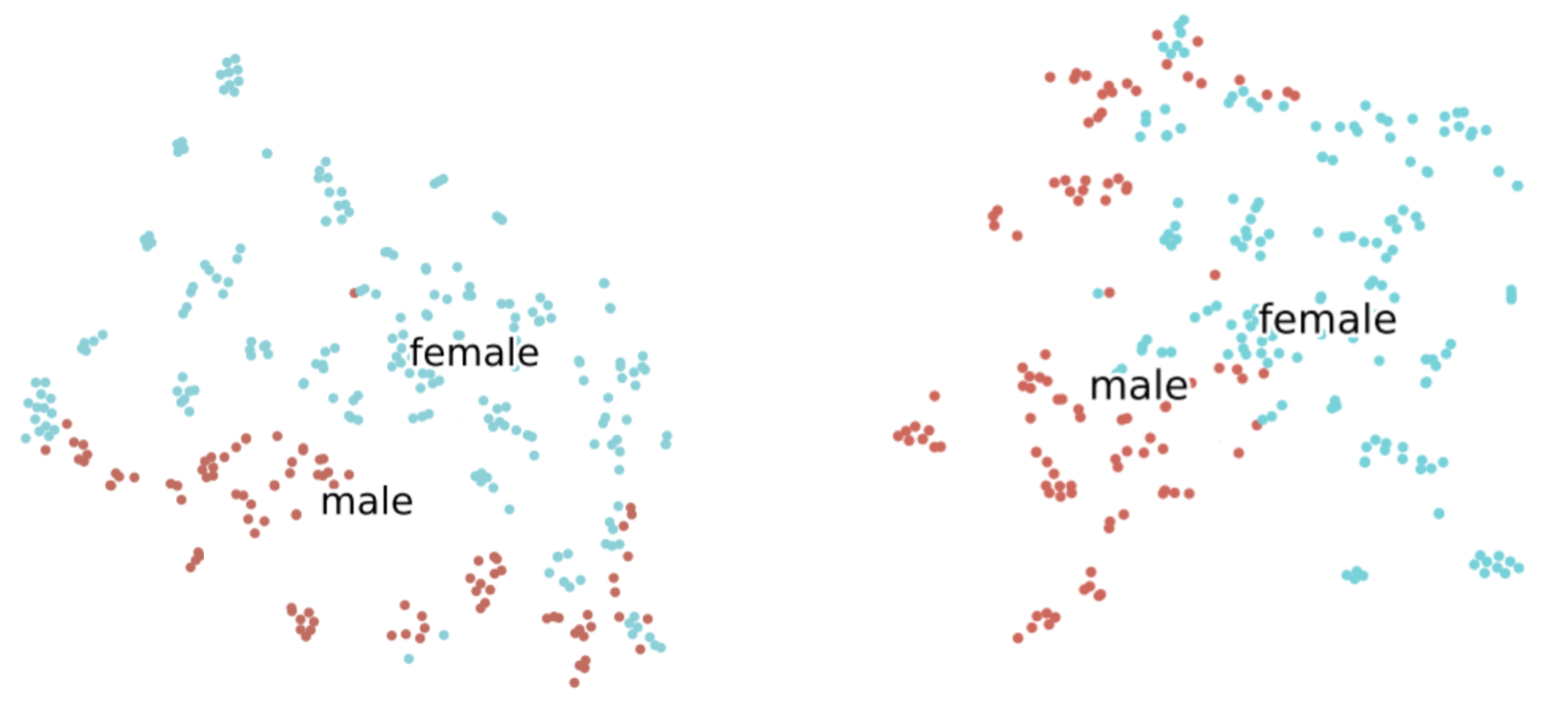} 
  \caption{Visualization of SCPM user embeddings for age group $30$ (left) and $44$ (right)  using T-SNE. } 
  \label{fig:user_emb_tsne2} 
\end{figure}

\textbf{Visualization of SCPM user embeddings.} We visualize the user embeddings from the SCPM model using T-SNE \cite{tsne2008}. Fig.~\ref{fig:user_emb_tsne2} shows the 2D projection of embeddings for users aged $30$ (left) and $44$ (right) from the Ponpare dataset, with colors indicating gender. The visualizations reveal distinct clustering patterns that suggest the SCPM model captures meaningful differences in user behavior across gender and age segments.

\section{Conclusion and Future Work} 
\label{sec:conclusion}

We present a unified deep learning framework for modeling and ranking heterogeneous treatment effects with complex real-world challenges. Our structured and continuous policy model enables arbitrary treatment representations beyond binary or discrete classes, and jointly optimizes treatment assignment, treatment intensity, with aggregation of multiple outcomes in the cost-aware objective function. Experimental results validate our approach, demonstrating improved performance in uplift ranking and cost-adjusted objective maximization. For future work, we intend to incorporate reinforcement learning techniques to dynamically adapt treatment policies based on sequential user interactions and long-term outcomes. This would enable the system to move beyond static treatment ranking to learning optimal policies over time. Moreover, incorporating proximal policy and advanced optimization techniques can benefit production models. Together, these directions will further integrate methodologies across causal inference, reinforcement learning, and business decision optimization, making them more effective in real-world deployment.


\bibliographystyle{ACM-Reference-Format}
\bibliography{references}
\appendix
\section{Constraint Ranking with Barrier Functions}
\label{constraint-ranking}
Given a batch of users, we create a barrier function that smoothly push effective probabilities towards zero when they don't meet constraints:
\begin{itemize}
    \item Sort the probabilities $\mathbf{p}=\{p_{1}, p_{2}, ..., p_{N}\}$ into $\{p_{k_1}, p_{k_2}, ..., p_{k_N}\}$, where each $k_j$ is a sorted index.
    \item Determine offset $d^*$ as a threshold\footnote{Or as a value in the in-between interval of two probabilities.} in the set $\mathbf{p}$ that defies the location of the barrier. 
    \item With $T$ as temperature hyperparameter to control the softness of the barrier, define the barrier function as: $\sigma(x) = \text{sigmoid}(-T (x - d^{*}))$.
    \item Multiply each user probability $p_{i}$ with its barrier function, i.e. $\hat{p}_{i} = p_{i}\sigma(p_{i})$. 
    This allows the algorithm to keep each $p_{i}$ if it is above the barrier \(d^{*}\) and nullify $p_{i}$ if it is below the barrier.
    \item Re-normalize $\hat{p}_{i}$ with softmax function so that they sum to one within the user cohort.
    \item Use the re-normalized probabilities to compute expectations in the $\tau$ functions.
\end{itemize}

\section{Propensity with Direct Ranking Model} \label{apB}
\subsection{Propensity Score}
The propensity score, denoted as $e(x)$, is the probability of an individual unit (a user in an experiment) receiving a treatment conditional on observed covariates $X$. Formally, it is defined as $e(\mathbf{x}) = P (T | \mathbf{X} = \mathbf{x})$ where T is a binary treatment indicator ($T=1$ for treatment, $T = 0$ for control). The propensity score is fundamental in causal inference, as it enables balancing covariates between treated and untreated groups to reduce confounding bias. By conditioning on $e(x)$, we can achieve a distribution of $X$ that is independent of treatment assignment, as established by \cite{Rubin1983}.

\subsection{Propensity Score Derivation for DRM}

As discussed in Eq. \ref{eq:pstatement_unconstrained_opt}, our aim is to minimize the cost per unit gain. For each user with feature $x$, $\tau^{*r}(\mathbf{x}) = E(Y_1^r - Y_0^r | \mathbf{X} = \mathbf{x})$ and cost $\tau^{*c}(\mathbf{x}) = E(Y_1^c - Y_0^c | \mathbf{X} = \mathbf{x})$ are expectations across selected user portfolios, where $Y_1^r$ and $Y_1^c$ are the reward and cost, respectively, for a treated user and $Y_0^r$ and $Y_0^c$ are the reward and cost for an untreated user.

Without loss of generality, we derive uplift $\tau$ without the subscript as it could be extended to $\tau^r$ and $\tau^c$ as below.
\ignorespacesafterend

We start with the fundamental definition of treatment effect:
\begin{align*}   
\tau&{*} = E(Y_1 - Y_0) = E(Y_1) - E(Y_0),
\end{align*}
where 
\begin{align*}
E(Y_1) &= E\left(\frac{Y_1}{e(\mathbf{x})} e(\mathbf{x})\right) \nonumber \\
       &= E\left(\frac{Y_1}{e(\mathbf{x})}E(T | \mathbf{X}=\mathbf{x})\right) \nonumber\\
       &= E\left(\frac{Y_1}{e(\mathbf{x})}E(T | \mathbf{X}=\mathbf{x}, Y_1)\right)\nonumber \\
       &= E\left(E\left(\frac{Y_1T}{e(\mathbf{x})} \middle| X, Y_1\right)\right) \nonumber\\
       &= E\left(\frac{Y_1T}{e(\mathbf{x})}\right),
\end{align*}
where the third step follows from unconfoundedness assumption~\cite{lunceford04stratification,nie2017quasi}. 

Similarly, $E(Y_0) = E(\frac{Y_0 (1-T)}{1-e(\mathbf{x})})$.  Therefore, combining the previous steps, we arrive at: \begin{align*} 
\begin{split} 
\tau^{*} &= E(Y_1 - Y_0) = E(Y_1) - E(Y_0) \\
&= E(\frac{Y_1T}{e(\mathbf{x})})  - E(\frac{Y_0 (1-T)}{1-e(\mathbf{x})}) \\ 
&= E(E(\frac{Y_1T}{e(\mathbf{x})}|T)) - E(E(\frac{Y_0(1-T)}{1-e(\mathbf{x})}|T)), \\ 
\end{split}
\end{align*} 
where the last step follows from the law of total expectation.

We can then expand the equation by definition of expectation: 
\begin{align*} 
    \begin{split} 
    \bar{\tau}^{*} &= P(T = 1) E(\frac{Y_1T}{e(\mathbf{x})}|T = 1) + \cancel{P(T = 0) E(\frac{Y_1T}{e(\mathbf{x})}| T = 0)}\\
    &\quad - \cancel{P(T = 1) E(\frac{Y_0 (1-T)}{1-e(\mathbf{x})}|T = 1)} - P(T = 0) E(\frac{Y_0 (1-T)}{1-e(\mathbf{x})} | T = 0) .
    \end{split} 
\end{align*} 
For each user, we use $p_i$ as a relaxation of the binary decision variable $z^{(i)}$ in the Duality R-Learner problem in Appendix \ref{dual_r_learner}, the effectiveness weighting to select user $i$. The higher the $p_i$, the higher likelihood of selecting the sample, and we constrain $\sum p_{i, T_i = 0} = 1$, $\sum p_{i, T_i=1} = 1$, i.e. the effectiveness measures for users in the treatment cohort, and control cohort sums to $1$ respectively. This normalization gives the effective measure a probabilistic interpretation. We define $\hat{e}$ to be the overall propensity, or probability of treatment, estimated by mean of the binary treatment labels across users. For each user, the $e(\mathbf{x})$ term is estimated from a pre-trained propensity function using logistic regression. Then, we expand the expectation terms $E(\frac{Y_1T}{e(\mathbf{x})}|T = 1)$, $E(\frac{Y_0 (1-T)}{1-e(\mathbf{x})} | T = 0)$ using the normalized effectiveness weighting, potential outcomes $Y^{(i)}_1$, $Y^{(i)}_0$ and observed outcome labels $Y^{(i)}$: 

\begin{align} 
\label{eq:drm_propensity_3} 
\begin{split} 
\tau^{*} &= \hat{e} \sum_{T_i =1} \frac{Y^{(i)}_1 p_{i}}{e(\mathbf{x}^{(i)})} - (1-\hat{e}) \sum_{T_i = 0} \frac{Y^{(i)}_0 p_i}{1-e(\mathbf{x}^{(i)})} \\ 
& = \hat{e} \sum_{i=1}^n \frac{1}{e(\mathbf{x}^{(i)})} Y^{(i)}_1 p_{i} \mathbb{I}_{T_i=1} - (1-\hat{e}) \sum_{i = 0}^n \frac{1}{1-e(\mathbf{x}^{(i)})} Y^{(i)}_0 p_i \mathbb{I}_{T_i=0} \\ 
& = \hat{e} \sum_{i=1}^n \frac{1}{e(\mathbf{x}^{(i)})} Y^{(i)} p_{i} \mathbb{I}_{T_i=1} - (1-\hat{e}) \sum_{i = 0}^n \frac{1}{1-e(\mathbf{x}^{(i)})} Y^{(i)} p_i \mathbb{I}_{T_i=0}.
\end{split} 
\end{align} 
Eq.~\ref{eq:drm_propensity_3} is a generalized form of the $\tau^{*}$ including part of the objective function with propensity weighting. When $\hat{e} = e(\mathbf{x}^{(i)})$ is constant in a fully randomized experiment, the term becomes: 
\begin{align*} 
  \bar\tau^{*}=\sum_{i=1}^n Y^{(i)}p_i(\mathbb{I}_{T_i=1} - \mathbb{I}_{T_i=0}).
\end{align*} 

For numerical stability and differentiability, we apply a rectified activation function $\sigma_r(\cdot)$ (such as softplus function), leading to the final form in Eq.~\ref{eq:propensity_drm_objective}.

\begin{equation}
\small
\label{eq:propensity_drm_objective}
\frac{\tau^{*c}}{\tau^{*r}} =
\frac{\sigma_r \left(\hat{e} \sum\limits_{i=1}^{n} \frac{1}{e(\mathbf{x}^{(i)})} Y^{c(i)} p_i \mathbb{I}_{T^{(i)}=1} - (1-\hat{e}) \sum\limits_{i=1}^{n} \frac{1}{1-e(\mathbf{x}^{(i)})} Y^{c(i)} p_i \mathbb{I}_{T^{(i)}=0} \right)}
{\sigma_r \left(\hat{e} \sum\limits_{i=1}^{n} \frac{1}{e(\mathbf{x}^{(i)})} Y^{r(i)} p_i \mathbb{I}_{T^{(i)}=1} - (1-\hat{e}) \sum\limits_{i=1}^{n} \frac{1}{1-e(\mathbf{x}^{(i)})} Y^{r(i)} p_i\mathbb{I}_{T^{(i)}=0} \right)}.
\end{equation}

This formulation enables counterfactual learning by integrating inverse propensity scoring into the optimization framework, ensuring an unbiased and stable estimation of aggregated effectiveness.

\section{Duality R-Learner Algorithm} \label{dual_r_learner}
We describe the duality method with Lagrangian multipliers to solve the constrained optimization problem for Eq. \ref{eq:constrained_problem}. 

Our objective is to maximize gain subject to a budget ($B>0$) constraint. To solve the problem, we relax $z^{(i)}\in\{0, 1\}$ variables to continuous ones, so the problem is defined as: 
\begin{align} 
\label{eq:constrained_problem} 
\begin{split}
  &\text{minimize} \quad -\sum_{i=1}^n\tau^{*r}(\mathbf{x}^{(i)}) \cdot z^{(i)} \\ 
  &\text{subject to} \quad \sum_{i=1}^n\tau^{*c}(\mathbf{x}^{(i)}) \cdot z^{(i)} \leq B \\
  & \text{where} \quad z^{(i)} \in\{0, 1\} \quad \text{relaxed to} \quad 0\leq z^{(i)} \leq 1.
\end{split}
\end{align} 
The variables $z^{(i)}$ represent whether we offer a reward to the user $i$ during a campaign and $B$ is the cost constraint.

We first fit the $\tau^*$ functions to the training data, by leveraging quasi-oracle estimation ~\cite{nie2017quasi}. 

Then, we have $\tau^{*r}(\mathbf{x}^{(i)})$ and $\tau^{*c}(\mathbf{x}^{(i)})$ given and fixed for each data point $\mathbf{x}^{(i)}$. We apply the Lagrangian multiplier \(\lambda\). The Lagrangian for problem \ref{eq:constrained_problem} is: 
\[
  L(\mathbf{z}, \lambda)=-\sum_{i=1}^n\tau^{*r}(\mathbf{x}^{(i)}) \cdot z^{(i)} + \lambda (\sum_{i=1}^n\tau^{*c}(\mathbf{x}^{(i)}) \cdot z^{(i)} - B). \\ 
\]
The optimization in problem~\ref{eq:constrained_problem} can then be rewritten in its Dual form to maximize the Lagrangian dual function $g = \inf_{\mathbf{z}\in\emph{D}}L(\mathbf{z}, \lambda)$: 
\begin{align}
\label{eq:duality_problem}
\max\limits_{\lambda} \inf_{\mathbf{z} \in \emph{D}} L(\mathbf{z}, \lambda) \quad
  \text{subject to} \ 0\leq z^{(i)} \leq 1, \lambda \geq 0.
\end{align}

To solve this problem using duality, we must consider certain caveats and determine whether the dual and primal problems achieve the same minimum.
\begin{itemize} 
\item Given $p(\mathbf{z}, \lambda) = -\sum_{i=1}^n\tau^{*r}(\mathbf{x}^{(i)}) \cdot z^{(i)}$, we know, for the optimal values of the primal and dual problems, the relationship $p^* \leq g^*$ holds according to the principles of convex optimization. Equality $p^* = g^*$ holds if $p$ and $g$ are convex functions, and the \emph{Slater constraint qualification} is met, which requires the primal problem to be strictly feasible. 
\item To satisfy the Slater condition, there must exist a feasible point such that the inequality constraints are strictly satisfied. For any positive value $B > 0$, we can always satisfy the strict inequality  $\sum_{i=1}^n\tau^{*c}(\mathbf{x}^{(i)}) \cdot z^{(i)} < B$ by selecting sufficiently small values for certain \(z^{(i)}\). Furthermore, in contexts like marketing campaign, the value of $B$ is usually large, which supports that the Slater qualifications hold in this setting. 
\end{itemize} 

\noindent From the analysis above, problem~\ref{eq:constrained_problem} and its dual problem~\ref{eq:duality_problem} are equivalent, and we can solve problem~\ref{eq:duality_problem} by iteratively optimizing with respect to $\mathbf{z}$ and $\lambda$.  

\textbf{Optimize $\mathbf{z^{(i)}}$:} Keeping $\lambda, \mathbf{\tau}$ fixed, as $\lambda$ and $B$ are constants, we can turn problem~\ref{eq:duality_problem} into: 
\begin{align*} 
\begin{split}
  \text{maximize}\quad \sum_{i=1}^nz^{(i)} s^{(i)} \\
  \text{subject to} \quad 0\leq z^{(i)} \leq 1,
\end{split}
\end{align*} 
where we define the \emph{effectiveness score} $s^{(i)} = \tau^{*r}(\mathbf{x}^{(i)})-\lambda\tau^{*c}(\mathbf{x}^{(i)})$. This optimization problem has a straightforward solution: assign the multiplier $z^{(i)} = 1$ when the ranking score $s^{(i)} \geq 0$; assign $z^{(i)} = 0$ when the ranking score $s^{(i)} < 0$. 

\textbf{Optimize $\lambda$:} We take the derivative of $L$ with regard to $\lambda$, $\frac{\partial g}{\partial \lambda}=B-\sum_{i=1}^n\tau^{*c}(\mathbf{x}^{(i)}) \cdot z^{(i)}$. We then update $\lambda$ by Eq. \ref{eq:update_lambda} where $\alpha$ is the learning rate. 
\begin{equation}
  \label{eq:update_lambda}
  \lambda\rightarrow \lambda + \alpha(B-\sum_{i=1}^n\tau^{*c}(\mathbf{x}^{(i)}) \cdot z^{(i)}).
\end{equation}
Based on the two steps above, we can iteratively solve for both $z^{(i)}$ and $\lambda$ \cite{bertsekas1999nonlinear} .

The above constrained optimization algorithm is implemented for our experiments. For a more simplified \emph{Duality R-learner} algorithm, we also implemented an approach to combine the two $\tau^*$ functions into one model. Instead of learning $\tau^{*r}$ and $\tau^{*c}$ respectively, we fit a single \emph{scoring model} $s^{(i)}=\tau^{*E}(\mathbf{x}^{(i)})$ in Eq. \ref{eq:lagrangian_score}. Note that the Duality solution suggests we should include any sample with $\hat\tau^{*E}(x^{(i)})>0$. The larger $\hat\tau^{*E}(x^{(i)})$ is, the more contribution the sample \(i\) will have and thus the higher ranking it should get. 

We define \(s^{(i)}\) to be:
\begin{equation} 
  \label{eq:lagrangian_score} 
  s^{(i)}=\tau^{*E}(\mathbf{x}^{(i)})=\tau^{*r}(\mathbf{x}^{(i)})-\lambda\tau^{*c}(\mathbf{x}^{(i)}) .
\end{equation} 
This form is linear, so we can use $Y^E=Y^r-\lambda Y^c$ instead of the the original $Y$ (single outcome for value and cost respectively) in the estimators above. Specifically, 
\begin{align*} 
  \tau^{*E} &= \tau^{*r}(\mathbf{x}) - \lambda \tau^{*c}(\mathbf{x}) \\
  &= E((Y^r_1 - \lambda Y^c_1 - (Y^r_0  - \lambda Y^c_0)| \mathbf{X} = \mathbf{x}) \\
  &= E(Y^E_1 - Y^E_0 | \mathbf{X} = \mathbf{x}) .
\end{align*} 
We then train the regression model through the quasi-oracle estimation method, and the output $\tau^{*E}$ could be used directly. This has two benefits: first, we optimize a joint model across $Y^r$ and $Y^c$ for the parameters to be able to find correlations jointly; second, for production and online service, we arrive at one single model to perform prediction. 

With the trained regression model, we are able to compute $\tau^{*r}$ and $\tau^{*c}$ for each user \(i\). We then re-compute \(z^{(i)}\) using the dual problem. This iterative process enables convergence toward an optimal solution for user selection and resource allocation.

\section{Model Configuration}
\label{model_config} 
We describe model implementation in detail. The source code is: \url{https://anonymous.4open.science/r/HCL-C9A0/README.md}.

\emph{SCPM}. We train SCPM with batch size being $8000$, with Adam optimizer of learning rate being $0.001$, running for $10$ epochs. The hidden dimension is  set to $32$ for all neural networks.

\emph{DragonNet and RERUM DragonNet}. We follow the architecture and training setup described in \cite{he2024rankability}.

\emph{CFR and RERUM CFR}. We follow the architecture and training setup described in \cite{he2024rankability}.

\emph{Quasi-oracle estimation (R-Learner)}. We use Linear Regression as the base estimator. Since we need to define one CATE function to rank users, we use the R-learner to model the gain value incrementality $\tau_r$. 

\emph{Causal Forest}. We leverage the generalized random forest (\emph{grf}) library in R~\cite{wager2018estimation,grflink,athey2016recursive}. To rank users with respect to cost vs. gain effectiveness, we estimate the conditional treatment effect function both for gain ($\tau_r$) and cost ($\tau_c$), i.e. train two Causal Forest models. For evaluation, we compute the ranking score according to the ratio of the two, i.e. $\frac{\tau_r}{\tau_c}$. For hyper-parameters, we perform search on deciles for parameters \emph{num\_trees, min.node.size}, and at \emph{0.05} intervals for \emph{alpha, sample.fraction parameters}. We also leverage the \emph{tune.parameters} option for the grf package. Best parameters are the same for all three datasets we experimented: \emph{num\_trees}$=100$ (50 trees for each of the two CATE function, $\tau_r$, $\tau_c$), \emph{alpha}$=0.2$, \emph{min.node.size}$=3$, \emph{sample.fraction}$=0.5$

\emph{R-Learner with Multi-layer Perceptron}. To study the effect of using a deep learning model with R-learner, we replace linear regression model in R-Learner with a two-layer neural network. With validation results, we find the optimal number of hidden layers to be $92$ for US Census and $100$ for Covertype data.

\emph{Duality R-Learner}. Similar to R-learner, we use Ridge Regression as the base estimator and constant propensity, and apply the model stated in Eq.~\ref{eq:lagrangian_score} for ease of online deployment. In practice, we select $\lambda$ with best performance on the validation set. We determine the value of $\lambda$ in the model  through hyper-parameter search on deciles and mid-deciles, e.g. $\lambda\in \{0.001, 0.005, 0.01, 0.05\}$; best $\lambda$ for marketing data is $0.1$, for US Census and Covertype data is $0.05$.

\emph{Direct Ranking}. We use a one layer network $\tanh(\mathbf{w}^T \mathbf{x} + b)$ as the scoring function without weight regularization, the objective function Eq.~\ref{eq:drm_deep_learning_summary}~\footnote{For numerical stability and differentiability, we apply a rectified activation function $\sigma(\cdot)$ (such as softplus function) to the denominator of Eq.~\ref{eq:drm_deep_learning_summary}.} is used. We implement the Adam optimizer with learning rate $0.001$, default beta values, run for $1500$ iterations.

\emph{Constrained Ranking}. We experiment with the Constrained Ranking Model with percentage constraints. We use a consistent percentage target at $40\%$. We apply a starting sigmoid temperature of $0.5$, and use annealing to increase temperature by $0.1$ every $10$ steps of Adam optimizer. The annealing schedule is obtained by cross-validation. 

\section{Datasets}
\label{appendix:data_sets}
\emph{US Census}. We select the treatment label as whether the person works more hours than the median of everyone else. The income (\emph{`dIncome1'}) is the gain dimension of outcome for $\tau^r$, and the number of children (`iFertil') multiplied by $-1.0$ is the cost dimension for estimating $\tau^c$.

\emph{Covtype}. Additionally, we would like to ensure the covertype trees are balanced by changing the hydrology with preference to `Spruce-Fir'. Thus, the treatment label is selected as whether the forest is close to hydrology, i.e., the distance to hydrology is below median of the filtered data. The gain outcome is a binary variable for whether distance to wild fire points is smaller than median, and cost outcome is the indicator for `Lodgepole Pine' ($1.0$, undesired) as opposed to `Spruce-Fir' ($0.0$, desired). 

\end{document}